\documentclass[runningheads]{llncs}
\usepackage{makeidx}
\usepackage{graphicx}
\usepackage{amsmath,amssymb} 
\usepackage{color}

\usepackage{amsfonts}
\usepackage{float}
\usepackage{tikz}
\usepackage{xcolor}
\usepackage{colortbl}
\usepackage{adjustbox}
\usepackage{breqn}	
\usepackage{soul}
\usepackage{framed}
\usepackage{multirow}
\usepackage{enumitem}
\usepackage{setspace}
\usepackage{hyperref}
\usepackage{makecell}
\usepackage{calc}
\usetikzlibrary{arrows}
\usetikzlibrary{shapes.geometric}
\usetikzlibrary{shapes.misc, positioning}
\usetikzlibrary{shapes.multipart, shapes.arrows}

\usepackage{xspace}
\newcommand{\eg}{e.g.\@\xspace}
\newcommand{\ie}{i.e.\@\xspace}

\newcommand{\cf}{cf.\@\xspace}
\newcommand{\etal}{et al.\@\xspace}

\tikzset{
	treenode/.style = {align=center, inner sep=0pt, text centered,
		font=\sffamily},
	tochild/.style={draw,-latex},
	toparent/.style={draw,latex-},
	noedge/.style={draw,latex-, white},
	blacknode/.style = {treenode, regular polygon,regular polygon sides=6, white, draw=black, fill=black, text width=1.5em}, 
	greynode/.style = {treenode, regular polygon,regular polygon sides=6, white, draw=black, fill=black!50, text width=1.5em}, 
	whitenode/.style = {treenode, regular polygon,regular polygon sides=6, black, draw=black, text width=1.5em, very thick}, 
	whitenodefill/.style = {treenode, circle, black, draw=black, fill=white, text width=1.5em}, 
	rednode/.style = {treenode, regular polygon,regular polygon sides=6, red, draw=red, text width=1.5em, very thick}, 
	fullrednode/.style = {treenode, regular polygon,regular polygon sides=6, white, draw=white, fill=red, text width=1.5em, very thick},
	block/.style= {draw, rectangle, minimum width=3cm,minimum height=1cm},
	smallblock/.style= {draw, rectangle, minimum width=2cm,minimum height=0.75cm},
}

\begin{document}
	\pagestyle{headings}
	\mainmatter

	\title{A Neural-Symbolic Architecture for Inverse Graphics Improved by Lifelong Meta-Learning}

	\titlerunning{Neural-Symbolic Architecture for Inverse Graphics}
	\authorrunning{Michael Kissner \and Helmut Mayer}
\author{
Michael Kissner \orcidID{0000-0002-8178-8236} \and
Helmut Mayer \orcidID{0000-0002-9439-2695}
}

\institute{
Institute for Applied Computer Science \\ Bundeswehr University Munich, Germany \\
\email{\{michael.kissner, helmut.mayer\}@unibw.de}
}

	\maketitle

	\begin{abstract}
	We follow the idea of formulating vision as inverse graphics and propose a new type of element for this task, a neural-symbolic capsule. It is capable of de-rendering a scene into semantic information feed-forward, as well as rendering it feed-backward. An initial set of capsules for graphical primitives is obtained from a generative grammar and connected into a full capsule network. Lifelong meta-learning continuously improves this network's detection capabilities by adding capsules for new and more complex objects it detects in a scene using few-shot learning. Preliminary results demonstrate the potential of our novel approach.
	\end{abstract}

\section{Introduction}

The idea of inverting grammar parse-trees to generate neural networks is not new \cite{Towell:1993,Towell:1994}, but has been largely abandoned. We revisit this idea and invert a generative grammar into a network of neural-symbolic capsules. Instead of labels, this capsule network outputs an entire scene-graph of the image, which is commonplace in modern game engines like Godot \cite{Team:2019}. Our approach is an inverse graphics pipeline for the prospective idea of an inverse game-engine \cite{Battaglia:2013}\cite{Ullman:2017}.

We begin by introducing the generative grammar and how to invert symbols and rules to obtain neural-symbolic capsules (Section 3). They are internally different to the ones proposed by Hinton \etal \cite{Hinton:2011}, as they essentially act as containers for regression models. Next, we present a modified routing-by-agreement and training protocol (Section 4), coupled to a lifelong meta-learning pipeline (Section 5). Through meta-learning, the capsule network continuously grows and trains individual capsules. We finally demonstrate the potential of the approach by presenting some results based on an "Asteroids"-like environment (Section 6) before ending with a conclusion.

\section{Related Work}\label{sec:RelatedWork}

\noindent\textbf{Capsule Networks.} In \cite{Hinton:2011} Hinton \etal introduced capsules, extending the idea of classical neurons by allowing them to output vectors instead of scalars. These vectors can be interpreted as attributes of an object and aim to reduce information-loss between layers of convolutional neural networks (CNN) \cite{Krizhevsky:2012,LeCun:1998}. Capsules require specialized routing protocols, such as routing-by-agreement \cite{Sabour:2017}, where the activation probability of a capsule is dependent on the agreement of its real inputs with their expected values. Further extensions for capsules have been proposed, such as using matrices internally \cite{Hinton:2018}, using 3D input \cite{Zhao:2018} and improving equivariance \cite{Lenssen:2018}.

\medskip

\noindent\textbf{Neural-Symbolic Methods.} There has been a strong effort to make a purely neural approach to vision more interpretable \cite{Lipton:2017,Mahendran:2015,Montavon:2018,Ribeiro:2016,Simonyan:2014,Zhang:2018}. An alternative approach to interpretability has been to deeply intertwine symbolic methods into connectionist methods. For computer vision, this is proving fruitful for de-rendering and scene decomposition  \cite{Kulkarni:2015,Liu:2019,Liu:2018,Tian:2019,Tulsiani:2017,Wu:2017,Zou:2017}. Many of these approaches use shape programs, a decomposition of the scene into a set of rendering instructions. The scene-graphs we construct can be viewed as such shape programs, but in a different representation. This symbolic information is well suited for more complex tasks, such as visual question answering (VQA) \cite{Mao:2019,Yi:2018} and scene manipulation \cite{Yao:2018}. We follow many of the presented ideas in our work.  

\section{Generative Grammar}

The neural-symbolic capsules are derived from a modified \cite{Martinovic:2013} generative attributed context-free grammar for image generation. We require that our grammar is non-recursive and has a finite number of symbols to avoid infinite productions. The following notation is used for our grammar $G$:

\begin{equation}
G = (S, V, \Sigma, R, A, C)
\end{equation}

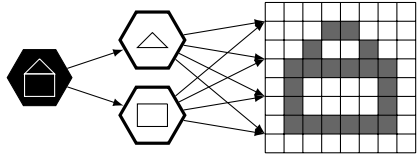
\begin{figure}[ht]
	\centering
	\begin{adjustbox}{max width=0.5\textwidth}
		\begin{tikzpicture}

		\node [blacknode] at (0, 0)  {}  
		child[grow=right]
		{ 
			node (A1) [whitenode] at (0, 0.5) {} edge from parent[tochild]
			{}
		}
		child[grow=right]
		{ 
			node (A2) [whitenode] at (0, -0.5) {} edge from parent[tochild]
			{}
		};
		
		\fill[black!60] (3.25,-0.75) rectangle (3.5,-0.5);
		\fill[black!60] (3.25,-0.5) rectangle (3.5,-0.25);
		\fill[black!60] (3.25,-0.25) rectangle (3.5,-0.0);
		\fill[black!60] (3.25,-0.0) rectangle (3.5,0.25);
		
		\fill[black!60] (4.5,-0.75) rectangle (4.75,-0.5);
		\fill[black!60] (4.5,-0.5) rectangle (4.75,-0.25);
		\fill[black!60] (4.5,-0.25) rectangle (4.75,-0.0);
		\fill[black!60] (4.5,-0.0) rectangle (4.75,0.25);
		
		\fill[black!60] (3.5,-0.0) rectangle (3.75,0.25);
		\fill[black!60] (3.75,-0.0) rectangle (4,0.25);
		\fill[black!60] (4.0,-0.0) rectangle (4.25,0.25);
		\fill[black!60] (4.25,-0.0) rectangle (4.5,0.25);
		
		\fill[black!60] (3.5,-0.75) rectangle (3.75,-0.5);
		\fill[black!60] (3.75,-0.75) rectangle (4,-0.5);
		\fill[black!60] (4.0,-0.75) rectangle (4.25,-0.5);
		\fill[black!60] (4.25,-0.75) rectangle (4.5,-0.5);
		
		\fill[black!60] (4.25,0.25) rectangle (4.5,0.5);
		\fill[black!60] (3.5,0.25) rectangle (3.75,0.5);
		\fill[black!60] (3.75,0.5) rectangle (4,0.75);
		\fill[black!60] (4.0,0.5) rectangle (4.25,0.75);
		
		\draw[scale=0.25] (12, -4) grid (20, 4);

		\draw[tochild, black] (A1) -- (3, 0.75);
		\draw[tochild, black] (A1) -- (3, 0.25);
		\draw[tochild, black] (A1) -- (3, -0.25);
		\draw[tochild, black] (A1) -- (3, -0.75);
		\draw[tochild, black] (A2) -- (3, 0.75);
		\draw[tochild, black] (A2) -- (3, 0.25);
		\draw[tochild, black] (A2) -- (3, -0.25);
		\draw[tochild, black] (A2) -- (3, -0.75);
		
		\draw [white] (0,0.25) -- (0.2,0.05) -- (-0.2,0.05) -- (0,0.25) ;
		\draw [white] (0.2, 0.05) -- (0.2,-0.25) -- (-0.2, -0.25) -- (-0.2, 0.05) -- (0.2, 0.05);
		
		\draw [black] (1.5,0.6) -- (1.7,0.4) -- (1.3,0.4) -- (1.5,0.6) ;
		\draw [black] (1.7, -0.35) -- (1.7,-0.65) -- (1.3, -0.65) -- (1.3, -0.35) -- (1.7, -0.35);
		
		\end{tikzpicture}
	\end{adjustbox}
	\caption{Draw: A grammar producing an image from a $\left[\textit{house\,}\right]$ symbol.} \label{fig:GrammarParseTree}
\end{figure}

\begin{itemize}[label={},itemindent=-1em,leftmargin=1em]
	\item\textbf{$S$:} The axiom (starting) symbol $S$ of our grammar is some object name (\eg, $\left[\textit{car\,}\right]$ or $\left[\textit{house\,}\right]$) for which we want to generate further, more detailed symbols.
	
	\item\textbf{$V$:} The finite set of non-terminal symbols $V$ represents parts of the axiom symbol (\eg, $\left[\textit{car door\,}\right]$) or parts of parts (\eg, $\left[\textit{door handle\,}\right]$). The further down the grammar parse-tree a symbol resides, the more primitive it becomes.
	
	\item\textbf{$\Sigma$:} The set of terminal symbols $\Sigma$ consists of graphical primitives. These may include elements such as $\left[\textit{edge\,}\right]$ or $\left[\textit{sphere\,}\right]$. Whichever terminal symbols are chosen will determine the possible complexity that can be represented by the non-terminal symbols.
	
	\item\textbf{$R$:} A production rule $r\in R$ is of the form
	
	\begin{equation}
	\Omega\to\lambda \;\;\;\; \text{with} \; \Omega\in V, \lambda\in \bigcup_{i\in\mathbb{N}\setminus\{0\}}(V\cup\Sigma) \;\;\;.
	\end{equation}
	The right-hand-side (RHS) of a rule $r$ has the form $\lambda = \lambda_1 \cdots \lambda_{\vert\lambda\vert}$, where $\lambda_i$ is either terminal or non-terminal and by $\vert\lambda\vert$ we denote the total number of produced symbols. There may be multiple rules in $R$ that have the same left-hand-side (LHS) symbol. We also introduce a special function called \textit{draw} that applies to terminal symbols, \ie, primitives, forcing them to produce a set of pixels corresponding to their graphical representation (\cf Figure \ref{fig:GrammarParseTree})
	
	\begin{equation}\label{eq:pixLayer}
	\textit{draw} : \lambda\to\left[\textit{pixel-layer}\right]\;\;\;\; \lambda\in\Sigma \;\;\;.
	\end{equation}

	\item\textbf{$A$:} Every terminal and non-terminal symbol $\lambda_j$ with rule $r$, as well as each $\left[\textit{pixel-layer\,}\right]$, is associated with an attribute vector $\vec{\alpha}_j = (\alpha_j^1, \cdots, \alpha_j^{n} )$.
	
	\item\textbf{$C$:} For $r$ to produce meaningful attributes, they must be constrained by a set of realistic laws that allow for a wide spectrum of results. Particularly, we introduce a set of non-linear equations which constrain the attributes. Each attribute $\alpha^i_j$ of a symbol $\lambda_j$ produced by rule $r : \Omega\to\lambda$ is associated with a constraint $g^i_j$ and calculated using $\alpha^i_j = g^i_j(\vec{\alpha}_\Omega)$. By $A(r)$ we denote the set of attributes $\{\vec{\alpha}_j\}$ and by $C(r)$ the set of all constraints $\{g^i_j\}$ for $r$. The \textit{draw} function is also considered to be such a constraint $g$.
	
\end{itemize}

The order of the symbols produced by a rule represents depth-sorting. For example, $\left[\textit{triangle}\right]\left[\textit{square\,}\right]$ first draws the triangle, then the square. We interpret two rules with the same LHS symbol to be equivalent to either drawing different unique \textbf{viewpoints} (from the front, from the back) of the same object, drawing a primitive using different part \textbf{configurations} (chair with padding, chair without padding) or drawing it in different \textbf{styles} (sketch, photo).

\section{Neural-Symbolic Capsules}\label{sec:Classes}

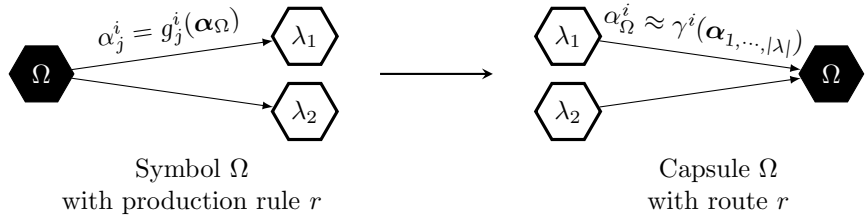
\begin{figure}[ht]
	\centering	
	\begin{adjustbox}{max width=1.0\textwidth}
		\begin{tikzpicture}
		
		\node [blacknode] at (0, 0)  {$\Omega$}  
		child[grow=right]{ node [whitenode] at (2, 0.5) {$\lambda_1$} edge from parent[tochild, sloped] node[above] {$\alpha_j^i=g^i_j(\vec{\alpha}_\Omega$)} }
		child[grow=right]{ node [whitenode] at (2, -0.5) {$\lambda_2$} edge from parent[tochild] };
		
		\draw (2, -1.5) node {\begin{tabular}{c} Symbol $\Omega$ \\ with production rule $r$ \end{tabular}};
		
		
		\draw [->, >=stealth, thick] (4.5,0) -- (6,0);
		
		\node [blacknode] at (10.5, 0)  {$\Omega$}  
		child[grow=left]{ node [whitenode] at (-2, 0.5) {$\lambda_1$} edge from parent[toparent, sloped] node [above]{$\alpha_\Omega^i\approx\gamma^i(\vec{\alpha}_{1,\cdots,\vert\lambda\vert})$} }
		child[grow=left]{ node [whitenode] at (-2, -0.5) {$\lambda_2$} edge from parent[toparent] };
		
		\draw (9, -1.5) node {\begin{tabular}{c} Capsule $\Omega$ \\ with route $r$ \end{tabular}};
		
		\end{tikzpicture}
	\end{adjustbox}
	\caption{Inversion of a symbol of the grammar parse-tree results in a capsule. We illustrate both the symbol and the capsule using a hexagon to avoid confusion with neurons.} \label{fig:reverseNode}
\end{figure}


In this section we introduce our neural-symbolic capsules. We "invert" a symbol of our grammar to create a capsule and connect it in reverse to the order in the parse-tree to form a capsule network. Our approach to the capsule's internals is different to \cite{Sabour:2017}. The idea of routing-by-agreement and vectorized outputs remains unchanged, however, we replace the internal algorithm by a regression model and output an additional activation probability. 

\medskip

\noindent\textbf{Terminal Symbols $\to$ Primitive Capsules.} Each terminal symbol represents a renderable graphical primitive that is connected directly to a layer of pixels and we refer to its inversion as a \textbf{primitive capsule}. These capsules perform detection based on pixel inputs. 

\medskip

\noindent\textbf{Non-Terminal Symbols $\to$ Semantic Capsules.} We invert non-terminal symbols to form \textbf{semantic capsules}. By $\Omega$ we henceforth interchangeably refer to both the capsule and its corresponding symbol.

\medskip

\noindent\textbf{Rules $\to$ Routes.} The constraints $g$ of a rule $r$ take the attributes $\vec{\alpha}_\Omega$ of symbol $\Omega$ and produce new attributes $\vec{\alpha}_1,\cdots\,\vec{\alpha}_{\vert\lambda\vert}$. After inversion, the capsule $\Omega$ takes those same attributes $\vec{\alpha}_1,\cdots\,\vec{\alpha}_{\vert\lambda\vert}$ as input to generate $\vec{\alpha}_\Omega$ using $g^{-1}$. However, $g$ is not invertible in most cases and we instead introduce $\gamma$ as our best approximation, such that

\begin{equation}\label{eq:Inv}
||g(\gamma(\vec{\alpha})) - \vec{\alpha}||
\end{equation}
is minimized (\cf Figure \ref{fig:reverseNode}). We refer to the inverted rule as a \textbf{route} and depending on context we denote by $r$ both the rule and the route. This also holds true for primitive capsules, where detection means the inversion of its \textit{draw} function.

\subsection{Routing-By-Agreement} 

We use a modified routing-by-agreement protocol to find the best fitting route and attributes. $\Omega$ may appear on the RHS of multiple routes (\ie, LHS of a rule), but as only one route leads to the activation of the capsule, we introduce an activation probability $p_r$ for each. Our goal is attribute equivariance and activation probability invariance under feature-preserving transformations. We propose the following internals of our capsule (\cf Figure \ref{fig:CapsuleBlockDiagram}):

\begin{figure}[ht]
	\centering
	\begin{adjustbox}{max width=1\textwidth}
		\begin{tikzpicture}
		
		\draw[draw=black!50, dashed, fill=black!5] (-30mm,40mm) rectangle (185mm, -72.5mm);
		\draw[draw=black!10!yellow, dashed, fill=yellow!10] (-22.5mm,30mm) rectangle (22.5mm, -62.5mm);
		\draw[draw=black!10!yellow, dashed, fill=yellow!10] (37.5mm,30mm) rectangle (82.5mm, -62.5mm);
		
		\node (InA) [whitenodefill] at (-37.5mm, 35mm) {$\vec{\alpha}_i$};
		\node (InP) [whitenodefill] at (-37.5mm, -67.5mm) {$p_i$};
		
		\node (A01) [trapezium, trapezium angle=-60, minimum height=15mm, draw=black!30!green, fill=black!10!green!20, thick] at (0, 15mm) {$\gamma_{r_1}$};
		\node (A02) [draw=red, fill=red!20, rounded rectangle, thick] at (0, 0) {$\vec{\alpha}_{\Omega_1}$};
		\node (A03) [trapezium, trapezium angle=60, minimum height=15mm, draw=black!30!green, fill=black!10!green!20, thick] at (0,-15mm) {$g_{r_1}$};
		\draw [tochild, black] (A01) -- (A02);
		\draw [tochild, black] (A02) -- (A03);
		\node (A0i) [rectangle, draw=black!30!green, text height=2.5mm, fill=black!10!green!20, thick] at (0, 22.5mm) {\footnotesize $\vec{\alpha}_i$};
		\node (A0o) [rectangle, draw=black!30!green, text height=2.5mm, fill=black!10!green!20, thick] at (0, -22.5mm) {\footnotesize $\vec{\tilde{\alpha}}_i$};
		
		\node (A0m) [rectangle, draw=black!30!blue, fill=black!10!blue!20, minimum height = 20mm, minimum width = 30mm, thick] at (0, -45mm) {Agreement};
		\node (A0mi1) [rectangle, draw=black!30!blue, text height=2.5mm, fill=black!10!blue!20, thick] at (0, -35mm) {\footnotesize $\vec{\tilde{\alpha}}_i$};
		\node (A0mi2) [rectangle, draw=black!30!blue, text height=2.5mm, fill=black!10!blue!20, thick] at (-15mm, -40mm) {\footnotesize $\vec{\alpha}_i$};
		\node (A0mi3) [rectangle, draw=black!30!blue, text height=2.5mm, fill=black!10!blue!20, thick] at (-15mm, -50mm) {\footnotesize $p_i$};
		\node (A0mo1) [rectangle, draw=black!30!blue, text height=2.5mm, fill=black!10!blue!20, thick] at (15mm, -45mm) {\footnotesize $p_{\Omega_1}$};
		\draw [tochild, black] (A0o) -- (A0mi1);
		
		\node (Ai1) at (27.5mm, 15mm) {$\cdots$};
		\node (Ai3) at (27.5mm, -15mm) {$\cdots$};
		
		\node (An1) [trapezium, trapezium angle=-60, minimum height=15mm, draw=black!30!green, fill=black!10!green!20, thick] at (60mm, 15mm) {$\gamma_{r_n}$};
		\node (An2) [draw=red, fill=red!20, rounded rectangle, thick] at (60mm, 0) {$\vec{\alpha}_{\Omega_n}$};
		\node (An3) [trapezium, trapezium angle=60, minimum height=15mm, draw=black!30!green, fill=black!10!green!20, thick] at (60mm,-15mm) {$g_{r_n}$};
		\draw [tochild, black] (An1) -- (An2);
		\draw [tochild, black] (An2) -- (An3);
		\node (Ani) [rectangle, draw=black!30!green, text height=2.5mm, fill=black!10!green!20, thick] at (60mm, 22.5mm) {\footnotesize $\vec{\alpha}_i$};
		\node (Ano) [rectangle, draw=black!30!green, text height=2.5mm, fill=black!10!green!20, thick] at (60mm, -22.5mm) {\footnotesize $\vec{\tilde{\alpha}}_i$};
		
		\node (Anm) [rectangle, draw=black!30!blue, fill=black!10!blue!20, minimum height = 20mm, minimum width = 30mm, thick] at (60mm, -45mm) {Agreement};
		\node (Anmi1) [rectangle, draw=black!30!blue, text height=2.5mm, fill=black!10!blue!20, thick] at (60mm, -35mm) {\footnotesize $\vec{\tilde{\alpha}}_i$};
		\node (Anmi2) [rectangle, draw=black!30!blue, text height=2.5mm, fill=black!10!blue!20, thick] at (45mm, -40mm) {\footnotesize $\vec{\alpha}_i$};
		\node (Anmi3) [rectangle, draw=black!30!blue, text height=2.5mm, fill=black!10!blue!20, thick] at (45mm, -50mm) {\footnotesize $p_i$};
		\node (Anmo1) [rectangle, draw=black!30!blue, text height=2.5mm, fill=black!10!blue!20, thick] at (75mm, -45mm) {\footnotesize $p_{\Omega_n}$};
		\draw [tochild, black] (Ano) -- (Anmi1);
		
		\draw[tochild, black] (InA) -- (0, 35mm) -- (A0i);
		\draw[tochild, black] (InA) -- (60mm, 35mm) -- (Ani);
		\draw[tochild, black] (InA) -- (-25mm, 35mm) -- (-25mm, -40mm) -- (A0mi2);
		\draw[tochild, black] (InP) -- (-25mm, -67.5mm) -- (-25mm, -50mm) -- (A0mi3);
		\draw[tochild, black] (InP) -- (35mm, -67.5mm) -- (35mm, -50mm) -- (Anmi3);
		\draw[tochild, black] (InA) -- (35mm, 35mm) -- (35mm, -40mm) -- (Anmi2);
		
		\node (OutABox) [rectangle, draw=black, text height=2.5mm, text width=30mm, thick, fill=white, minimum height = 20mm, minimum width = 30mm, align=center] at (120mm, -5mm) {Route\\Selection};
		\node (OutAi1) [rectangle, draw=black, fill=white, text height=2.5mm, thick] at (105mm, 0) {\footnotesize $\vec{\alpha}_{\Omega_j}$};
		\node (OutAi2) [rectangle, draw=black, fill=white, text height=2.5mm, thick] at (105mm, -10mm) {\footnotesize $p_{\Omega_j}$};
		\node (OutAo1) [rectangle, draw=black, fill=white, text height=2.5mm, thick] at (135mm, 0mm) {\footnotesize $\vec{\alpha}_\Omega$};
		\node (OutPo1) [rectangle, draw=black, fill=white, text height=2.5mm, thick] at (135mm, -10mm) {\footnotesize $p_\Omega$};
		
		\node (OutA) [whitenodefill] at (195mm, -0mm) {$\vec{\alpha}_\Omega$};
		\node (OutP) [whitenodefill] at (195mm, -10mm) {$p_\Omega$};
		
		\draw[tochild, blue, line width=1mm] (Anmo1) -- (90mm,-45mm) -- (90mm,-10mm) -- (OutAi2);
		\draw[blue, line width=1mm] (A0mo1) -- (Anm);
		\draw[tochild, red, line width=1mm] (An2) -- (OutAi1);
		\draw[red, line width=1mm] (A02) -- (An2);
		\draw[tochild, black] (150mm, -0mm) -- (OutA);
		\draw[tochild, black] (150mm, -10mm) -- (OutP);
		\draw[tochild, black] (OutAo1) -- (150mm, -0mm);
		\draw[tochild, black] (OutPo1) -- (150mm, -10mm);
		
		\node[text=black!50] at (177mm, 35.5mm) {Capsule};
		\node[text=black!50!yellow] at (14.5mm, -60.5mm) {Route 1};
		\node[text=black!50!yellow] at (74.5mm, -60.5mm) {Route n};
		
		
		\draw[draw=black, fill=white, thick, dashed] (145mm, -15mm) rectangle (180mm, -62.5mm);
		\draw[draw=black, fill=white, thick] (145mm, 5mm) rectangle (180mm, -15mm);
		\node[text=black, text width=30mm, align=center] at (162.5mm, -5mm) {Observation\\Table};
		\node[text=black] at (162.5mm, -19mm) {$(p_\Omega, \vec{\alpha}_\lambda, \vec{\alpha}_\Omega)^{(1)}$};
		\node[text=black] at (162.5mm, -27mm) {$(p_\Omega, \vec{\alpha}_\lambda, \vec{\alpha}_\Omega)^{(2)}$};
		\node[text=black] at (162.5mm, -35mm) {$(p_\Omega, \vec{\alpha}_\lambda, \vec{\alpha}_\Omega)^{(3)}$};
		\node[text=black] at (162.5mm, -43mm) {\textbf{$\vdots$}};
		\draw[draw=black, dashed] (145mm, -23mm) -- (180mm, -23mm);
		\draw[draw=black, dashed] (145mm, -31mm) -- (180mm, -31mm);
		\draw[draw=black, dashed] (145mm, -39mm) -- (180mm, -39mm);
		
		\node (InAL) [rectangle, draw=black, fill=white, text height=2.5mm, thick] at (145mm, 0mm) {\footnotesize $\vec{\alpha}_\Omega$};
		\node (InPL) [rectangle, draw=black, fill=white, text height=2.5mm, thick] at (145mm, -10mm) {\footnotesize $p_\Omega$};
		\node (OutAL) [rectangle, draw=black, fill=white, text height=2.5mm, thick] at (180mm, 0mm) {\footnotesize $\vec{\alpha}_\Omega$};
		\node (OutPL) [rectangle, draw=black, fill=white, text height=2.5mm, thick] at (180mm, -10mm) {\footnotesize $p_\Omega$};
		
		\end{tikzpicture}
	\end{adjustbox}
	\caption{The inner structure of a capsule $\Omega$ with inputs $\vec{\alpha}_i, p_i$ and outputs $\vec{\alpha}_\Omega, p_\Omega$, representing our routing-by-agreement protocol (Equations \ref{eq:CapsuleInternals0} to \ref{eq:CapsuleInternals4}). The outputs are stored in an observation table and individual routes are highlighted in yellow. } \label{fig:CapsuleBlockDiagram}
\end{figure}
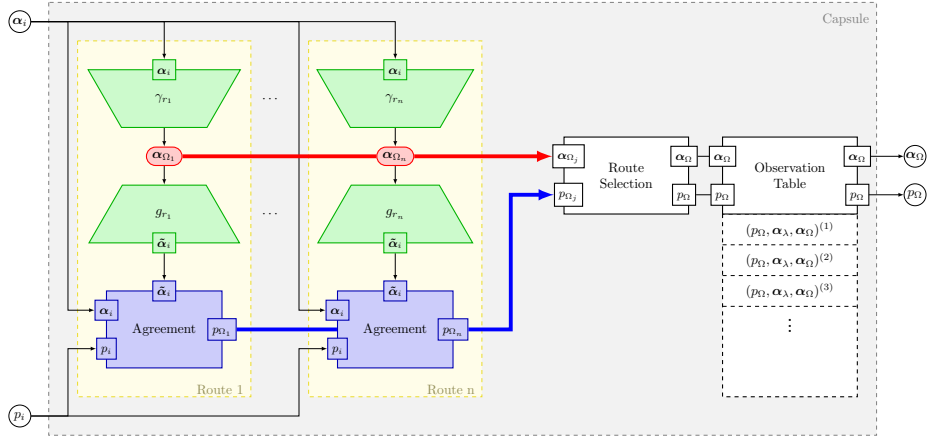

\begin{enumerate}
	
	\item The output $\vec{\alpha}_{\Omega_r}$ for a route $r$ is calculated using
	
	\begin{equation}\label{eq:CapsuleInternals0}
	\vec{\alpha}_{\Omega_r}=\gamma_r(\vec{\alpha}_{1,\cdots,\vert\lambda\vert})\;\;\; .
	\end{equation}
	
	\item For each input $\vec{\alpha}_{j_r}$ for a route $r$, we estimate the expected input value $\vec{\tilde{\alpha}}_{j_r}$ as if $\vec{\alpha}_{j_r}$ were unknown, using the following equation:
	
	\begin{equation}\label{eq:CapsuleInternals1}
	\vec{\tilde{\alpha}}_{j} = g_{r,j}(\vec{\alpha}_{\Omega_r})\;\;\; .
	\end{equation}
	
	\item The activation probability $p_{\Omega_r}$ of a route $r$ is calculated as 
	
	\begin{equation}\label{eq:CapsuleInternals2}
	p_{\Omega_r} = \frac{1}{\vert(\lambda)_r\vert}\;\;\sum_{(\lambda)_r}\; \frac{\lVert Z\left(\vec{\alpha}_i, \vec{\tilde{\alpha}}_i\right)\rVert_1}{\vert Z \vert} \cdot w\left(\frac{p_{i}}{\bar{p}_i} - 1 \right)\;\;\; ,
	\end{equation}
	where $(\lambda)_r$ denotes the set of all inputs that contribute to a route $r$, $p_{i}$ the route's input capsule's probability of activation, $Z$ an agreement-function with output vector of size $\vert Z \vert$, $\lVert \cdot \rVert_1$ the $l_1$-norm, $w$ some window function with $w(0) = 1$, $\sup\{w\} = 1$ and $\bar{p}_i$ the past mean probability for that input. 
	
	\item Steps 1. - 3. are repeated for each $r\in R(\Omega)$.
	
	\item Find the route that was most likely used
	
	\begin{equation}\label{eq:CapsuleInternals3}
	r_{\textit{final}} = \max_r \{ p_{\Omega_r} \}
	\end{equation}
	and set the final output as
	
	\begin{align}\label{eq:CapsuleInternals4}
	&p_\Omega = p_{\Omega_{r_{\textit{final}}}} \\
	&\vec{\alpha}_\Omega = \vec{\alpha}_{\Omega_{r_{\textit{final}}}} \;\;\; . 
	\end{align}
	
\end{enumerate}

Steps 1 and 2 correspond to an architecture equivalent to a de-rendering autoencoder, $g(\gamma(\vec{\alpha})) = \vec{\tilde{\alpha}}$, but with known interpretation for the latent variables (attributes). Here, $\gamma$ acts as encoder and $g$ as decoder.

For now assume that $\bar{p}_i$ is known in step 3. The agreement-function $Z$ measures how well the inputs of $\gamma$ correspond to the outputs of $g$. For semantic capsules, we choose the agreement-function

\begin{equation}\label{eq:SemantsicAgreementFunction}
Z\left(\vec{\alpha}_i, \vec{\tilde{\alpha}}_i\right) = \max\{ w\left(\vec{\tilde{\alpha}}_{i} - \vec{\hat{\alpha}}_{i} \right) \; : \; \vec{\hat{\alpha}}_{i} \in R_{\vec{\alpha}_i} \} \;\;\;,
\end{equation}
where $w$ describes an $n$-dimensional window function and $R_{\vec{\alpha}_i}$ the set of rotationally equivalent $\vec{\alpha}_i$. For primitive capsules finding an appropriate $Z$ depends very much on the design and symmetries of the decoder $g$ (\textit{draw}-function).

\subsection{Connecting Capsules}\label{sec:ConnectingCapsules}

Individual capsules are connected as shown in Figure \ref{fig:reverseNode}. A full capsule network is constructed from multiple grammars with different axioms ($\left[\textit{table\,}\right]$, $\left[\textit{chair\,}\right]$, ...). To avoid multiple capsules for the same symbol in the network, we merge them and introduce an observation table $\Lambda$ that stores all occurrences on a per-image basis (\cf Figure \ref{fig:CapsuleBlockDiagram}). For instance, a $\left[\textit{table}\right]$ capsule does not need to connect to four $\left[\textit{table-leg\,}\right]$ capsules, but only to one with four entries in its observation table. 

As these observations are reset at the beginning of each pass, we assume that all past entries in $\Lambda$ are stored in permanent memory elsewhere as

\begin{equation}\label{eq:memoryTrainingSet}
\left((p_\lambda)^{(i)}, (\vec{\alpha}_\lambda)^{(i)},(\vec{\alpha}_\Omega)^{(i)}\right)_r \;\;\; ,
\end{equation} 
allowing us to calculate the mean value $\bar{p}_\lambda$, as well as to perform meta-learning.

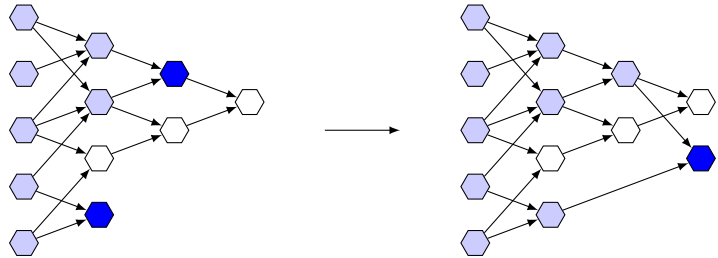
\begin{figure}[ht]
	\centering
	\begin{adjustbox}{max width=1.0\textwidth}
		\begin{tikzpicture}
		
		\node (C01) [regular polygon,regular polygon sides=6, draw=black, fill=blue!20!white] at (0,-1.5) {};
		\node (C02) [regular polygon,regular polygon sides=6, draw=black, fill=blue!20!white] at (0,-0.75) {};
		\node (C03) [regular polygon,regular polygon sides=6, draw=black, fill=blue!20!white] at (0,0) {};
		\node (C04) [regular polygon,regular polygon sides=6, draw=black, fill=blue!20!white] at (0,0.75) {};
		\node (C05) [regular polygon,regular polygon sides=6, draw=black, fill=blue!20!white] at (0,1.5) {};
		
		\node (C11) [regular polygon,regular polygon sides=6, draw=black, fill=blue] at (1,-1.125) {};
		\node (C12) [regular polygon,regular polygon sides=6, draw=black, fill=white] at (1,-0.375) {};
		\node (C13) [regular polygon,regular polygon sides=6, draw=black, fill=blue!20!white] at (1,0.375) {};
		\node (C14) [regular polygon,regular polygon sides=6, draw=black, fill=blue!20!white] at (1,1.125) {};
		
		\node (C22) [regular polygon,regular polygon sides=6, draw=black, fill=white] at (2,0.0) {};
		\node (C23) [regular polygon,regular polygon sides=6, draw=black, fill=blue] at (2,0.75) {};
		
		\node (C31) [regular polygon,regular polygon sides=6, draw=black, fill=white] at (3,0.375) {};
		
		\draw[tochild, black] (C01) -- (C11);
		\draw[tochild, black] (C01) -- (C12);
		\draw[tochild, black] (C02) -- (C11);
		\draw[tochild, black] (C02) -- (C13);
		\draw[tochild, black] (C03) -- (C13);
		\draw[tochild, black] (C03) -- (C14);
		\draw[tochild, black] (C04) -- (C14);
		\draw[tochild, black] (C05) -- (C14);
		\draw[tochild, black] (C05) -- (C13);
		\draw[tochild, black] (C03) -- (C12);
		
		\draw[tochild, black] (C12) -- (C22);
		\draw[tochild, black] (C13) -- (C22);
		\draw[tochild, black] (C13) -- (C23);
		\draw[tochild, black] (C14) -- (C23);
		
		\draw[tochild, black] (C22) -- (C31);
		\draw[tochild, black] (C23) -- (C31);
		
		
		\draw[tochild, black] (4,0) -- (5,0);
		
		\node (C01) [regular polygon,regular polygon sides=6, draw=black, fill=blue!20!white] at (6,-1.5) {};
		\node (C02) [regular polygon,regular polygon sides=6, draw=black, fill=blue!20!white] at (6,-0.75) {};
		\node (C03) [regular polygon,regular polygon sides=6, draw=black, fill=blue!20!white] at (6,0) {};
		\node (C04) [regular polygon,regular polygon sides=6, draw=black, fill=blue!20!white] at (6,0.75) {};
		\node (C05) [regular polygon,regular polygon sides=6, draw=black, fill=blue!20!white] at (6,1.5) {};
		
		\node (C11) [regular polygon,regular polygon sides=6, draw=black, fill=blue!20!white] at (7,-1.125) {};
		\node (C12) [regular polygon,regular polygon sides=6, draw=black, fill=white] at (7,-0.375) {};
		\node (C13) [regular polygon,regular polygon sides=6, draw=black, fill=blue!20!white] at (7,0.375) {};
		\node (C14) [regular polygon,regular polygon sides=6, draw=black, fill=blue!20!white] at (7,1.125) {};
		
		\node (C22) [regular polygon,regular polygon sides=6, draw=black, fill=white] at (8,0.0) {};
		\node (C23) [regular polygon,regular polygon sides=6, draw=black, fill=blue!20!white] at (8,0.75) {};
		
		\node (C31) [regular polygon,regular polygon sides=6, draw=black, fill=white] at (9,0.375) {};
		\node (C32) [regular polygon,regular polygon sides=6, draw=black, fill=blue] at (9,-0.375) {};
		
		\draw[tochild, black] (C01) -- (C11);
		\draw[tochild, black] (C01) -- (C12);
		\draw[tochild, black] (C02) -- (C11);
		\draw[tochild, black] (C02) -- (C13);
		\draw[tochild, black] (C03) -- (C13);
		\draw[tochild, black] (C03) -- (C14);
		\draw[tochild, black] (C04) -- (C14);
		\draw[tochild, black] (C05) -- (C14);
		\draw[tochild, black] (C05) -- (C13);
		\draw[tochild, black] (C03) -- (C12);
		
		\draw[tochild, black] (C12) -- (C22);
		\draw[tochild, black] (C13) -- (C22);
		\draw[tochild, black] (C13) -- (C23);
		\draw[tochild, black] (C14) -- (C23);
		
		\draw[tochild, black] (C22) -- (C31);
		\draw[tochild, black] (C23) -- (C31);
		
		\draw[tochild, black] (C11) -- (C32); 
		\draw[tochild, black] (C23) -- (C32);
		\end{tikzpicture}
	\end{adjustbox}
	\caption{A capsule network with all activated capsules in blue (left). Here, the topmost activated capsules (dark blue) do not have a common parent capsule that activated. In this case the meta-learning agent adds a common parent / axiom (right).} \label{fig:DanglingCaps}
\end{figure}

During a forward pass the entries in all the observation tables form one or many tree structures, which we call the \textbf{observed parse-trees}. Their topmost symbol is not necessarily one of the axioms of the grammars the capsule network is based on (\cf left of Figure \ref{fig:DanglingCaps}). Each observed parse-tree, thus, induces its own \textbf{observed grammar} with the topmost symbol being its \textbf{observed axiom}.

For the multitude of observed grammars in the capsule network, we postulate that we can always define a higher-level grammar by simply taking their union and defining a new axiom with a rule that produces the previous axioms (\cf Figure \ref{fig:DanglingCaps}). For example, the grammars for $\left[\textit{table\,}\right]$ and $\left[\textit{chair\,}\right]$ allow us to define a meaningful higher-level grammar with $\left[\textit{dining-room\,}\right]$ as the axiom. 

\subsection{Training Capsules}\label{sec:knownGamma}

\noindent\textbf{Training Primitive Capsules.} We assume the decoder $g$ (\textit{draw}) for primitive capsules is known. Finding an analytical solution to $\gamma$ is out of reach. Instead, we use a regression model for $\gamma$ and synthesize training sets with $g$. We define the inputs to $g$ using quantile functions $Q_j(p) = p$ for each attribute $\alpha^j_\Omega$, which we may refine according to our prior knowledge. Next, with $\chi_{i,j}$ some uniform random variable, $\chi_{i,j} \sim U( [0,1])$ and $f$ some function that applies random backgrounds, occlusions and special effects, we generate $\gamma$'s virtually infinite training set using

\begin{equation}
\left((f(g(Q_j(\chi_{i,j}))))^{(i)}, (Q_j(\chi_{i,j}))^{(i)} \right)\;\;\;.
\end{equation}

\medskip

\noindent\textbf{Training Semantic Capsules.} If only $\gamma$ of a route is known and $g$ unknown, we use a similar method to the case above. Ideally, we calculate $\alpha_\Omega$ using $\gamma$ and train $g$ using the training sets

\begin{equation}\label{eq:OriginalTrainingSet}
((\gamma(\vec{\alpha}_\lambda))^{(i)}, (\vec{\alpha}_\lambda)^{(i)}) \;\;\; .
\end{equation}

We must, however, first find a suitable $\gamma$. The initial output attributes of our semantic capsules consist of the distinct set of all input attributes $\bigcup_i A(r_i)$. We are free in our choice for $\gamma$, which is non-injective in most cases. It is expected that there will be collisions, \ie, different sets of inputs leading to the same output. These collisions are the main focus of our meta-learning pipeline. To minimize these collisions, we choose a $\gamma$ that calculates the mean of inputs of the same type, weighted by their size (width, height, depth). This weighting by size is to ensure that, for example, a wooden chair with many metallic screws is still considered wooden instead of metallic. For a general $k$th attribute we have:

\begin{equation}\label{eq:Gamma1}
\vec{\alpha}^{\; k}_\Omega = \gamma^{\; k}(\vec{\alpha}_\lambda) = \frac{1}{\sum_\lambda \Vert\vec{\alpha}^{\; size}_\lambda\Vert} \sum_\lambda \vec{\alpha}^{\; k}_\lambda \cdot \Vert\vec{\alpha}^{\; size}_\lambda\Vert   \;\;\; .
\end{equation}
However, we use special functions for size and position

\begin{equation}
\begin{split}
\vec{\alpha}^{\; size}_\Omega = \gamma^{\; size}(\vec{\alpha}_\lambda) =  & \max_{\lambda, i} \left( \textbf{R}^{-1}_\Omega \cdot (\vec{\alpha}^{\; pos}_\lambda +  \textbf{R}_\lambda\cdot \vec{B}_{\lambda,i}) \right)  \\ 
& - \min_{\lambda, i} \left( \textbf{R}^{-1}_\Omega \cdot (\vec{\alpha}^{\; pos}_\lambda +  \textbf{R}_\lambda\cdot \vec{B}_{\lambda,i}) \right)
\end{split}
\end{equation}
\begin{equation}\label{eq:Gamma3}
\begin{split}
\vec{\alpha}^{\; pos}_\Omega = \gamma^{\; pos}(\vec{\alpha}_\lambda) = \textbf{R}_\Omega \cdot \frac{1}{2}  & \left[ \max_{\lambda, i} \left( \textbf{R}^{-1}_\Omega \cdot (\vec{\alpha}^{\; pos}_\lambda +  \textbf{R}_\lambda\cdot \vec{B}_{\lambda,i}) \right) \right. \\ 
& \left. + \min_{\lambda, i} \left( \textbf{R}^{-1}_\Omega \cdot (\vec{\alpha}^{\; pos}_\lambda +  \textbf{R}_\lambda\cdot \vec{B}_{\lambda,i}) \right) \right]  \;\;\;,
\end{split}
\end{equation}
to ensure that they are in the correct reference frame. Here, $\vec{\alpha}^{\; rot}, \vec{\alpha}^{\; size}, \vec{\alpha}^{\; pos}$ are the vectorized subsets of the attribute vector $\vec{\alpha}$ for rotation, size and position, $\textbf{R}_\lambda$ and $\textbf{R}_\Omega$ indicate the Euler rotation matrix calculated from the rotation attributes $\vec{\alpha}_\lambda^{rot}$ and $\vec{\alpha}_\Omega^{rot}$ and $\vec{B}_{\lambda, i}$ the $i$th corner position vector of the bounding box of $\lambda$ (\ie, pairwise permutations of $\vec{\alpha}^{\; size}_\lambda / 2$ and $-\vec{\alpha}^{\; size}_\lambda / 2$). 

We can't create arbitrary inputs for training. Instead we use observations from memory (\cf Equation \ref{eq:memoryTrainingSet}) for augmentation. Detection needs to be invariant under changes of the outputs $\vec{\alpha}^{\; pos}_\Omega$, $\vec{\alpha}^{\; rot}_\Omega$ and $\vec{\alpha}^{\; size}_\Omega > 0$ and we let $T_i(\cdot)$ denote transformation functions that rotate (acting on $\vec{\alpha}^{\; pos}_\lambda$, $\vec{\alpha}^{\; rot}_\lambda$), translate (acting on $\vec{\alpha}^{\; pos}_\lambda$) and scale (acting on $\vec{\alpha}^{\; pos}_\lambda$, $\vec{\alpha}^{\; size}_\lambda$) all parts, while leaving the relative rotation, position and size to each other unchanged.

For the other attributes $\vec{\alpha}^k_\lambda$ we have little knowledge on how to perform feature-preserving transformations. However, if our original training set (\cf Equation \ref{eq:OriginalTrainingSet}) contains an output attribute $\vec{\alpha}^k_\Omega$ for which all entries are smaller than some $\epsilon$, we can safely assume that we have never encountered an object with this attribute and are free to "invent" possible values for this attribute type by simply setting all input $\vec{\alpha}^k_\lambda$ uniformly to some constant. For example, if our training set is filled with real apples, for which the stem is brown and the body red or green, we can invent a metallic apple by simply assuming that both the stem and body are metallic, as we have no idea what it really would look like. By $U_i(\cdot)$ we denote a linear "style" transformation that sets a constant value in the range $[0,1]$ to all unused attributes of the same type and either activates or deactivates it. We finally have our fully augmented set for training $g$:

\begin{equation}\label{eq:AugmentedTrainingSet}
((\gamma(T_i(U_i(\vec{\alpha}_\lambda))))^{(i)}, (T_i(U_i(\vec{\alpha}_\lambda)))^{(i)}) \;\;\; .
\end{equation}

A single example is sufficient for the above training regime to begin augmentation by translating, rotating and resizing the object ($T_i$), as well as trying out different styles ($U_i$). 

\medskip

\noindent\textbf{New Attributes and Re-Training.} For semantic capsules, the set of attributes is not static and can grow. We differentiate between adding attributes due to inheritance and due to a trigger from the meta-learning agent.

Inheritance occurs automatically when one of the input capsules is extended by an attribute that is unknown to the current capsule. We simply expand the capsule's internals by said attribute and retrain it. This inheritance propagates down the network, forcing subsequent capsules to inherit them as well.

The more interesting case arises when the capsule is triggered by meta-learning to expand its attributes by adding $\alpha^{\textit{new}}_\Omega$. First, we expand every attribute vector in memory for this capsule by the new attribute, but set to $\alpha^{\textit{new}}_\Omega = 0$. Next, the internal attribute vector of the capsule and its $\gamma$ and $g$ functions are extended. We begin by replacing $g$ with a new regression model of increased input width. The problem here is that we require $\gamma$ to train it, which at this point has not been extended yet. Instead we split $\vec{\alpha}$ and $\gamma$ into two parts, one containing the newly added attribute $\alpha^{\textit{new}}_\Omega = \gamma^{\textit{new}}(\vec{\alpha}_\lambda)$ as an output and one with the previous attributes as output $\vec{\alpha}^{\textit{old}}_\Omega = \gamma^{\textit{old}}(\vec{\alpha}_\lambda)$:

\begin{equation}
\gamma(\vec{\alpha}_\lambda) = \left(\gamma^{\textit{old}}(\vec{\alpha}_\lambda) \;\oplus\; \gamma^{\textit{new}}(\vec{\alpha}_\lambda) \right)\;\;\;,
\end{equation}
where by $\vec{a} \oplus \vec{b}$ we mean the concatenation of two vectors. At this point $\gamma^{\textit{old}}$, $\alpha^{\textit{new}}_\Omega$ and $\vec{\alpha}_\lambda$ are known and this suffices to start training $g$ using

\begin{equation}
\bigg((\gamma^{\textit{old}}(T_i(U_i(\vec{\alpha}_\lambda))) \;\oplus\; \alpha^{\textit{new}}_\Omega)^{(i)}, (T_i(U_i(\vec{\alpha}_\lambda)))^{(i)}\bigg) \;\;\; .
\end{equation}

Finally we need to determine $\gamma^{\textit{new}}$. We add a regression model with one output that runs in parallel to $\gamma^{\textit{old}}$ and train it using the new decoder $g$:

\begin{equation}
\bigg((g(\vec{\alpha}^{\textit{old}}_\Omega  \;\oplus\; \alpha^{\textit{new}}_\Omega))^{(i)}, (\alpha^{\textit{new}}_\Omega)^{(i)} \bigg)\;\;\; .
\end{equation}

\section{Meta-Learning}

It is far too difficult to define the entire grammar with all rules and constraints from scratch to generate a complete capsule network. Instead, our approach works bottom-up and we only define the terminal symbols (primitive capsules), letting the meta-learning agent learn all semantic capsules and routes. This means that our initial set of primitive capsules limits what the network can eventually analyze and learn. Ideally, we would define primitive capsules for the most basic set of primitives from which we are able to construct every kind of object. We can, however, refine this set later on. A grammar with $\left[\textit{square\,}\right]$ as terminal symbol produces the same results, even if it is refined by $\left[\textit{edge\,}\right]$ terminal symbols with the rule $ \left[\textit{square}\right]\to\left[\textit{edge}\right]\left[\textit{edge}\right]\left[\textit{edge}\right]\left[\textit{edge\,}\right]$. For the \textit{draw} functions, we rely on the current state of computer graphics. Here we have access to a near endless supply of parameterizable primitives \cite{Quílez:2017}\cite{Zhou:2018} and graphics pipelines capable of physically-based rendering \cite{Pharr:2016}. By Equation \ref{eq:Inv}, if we can render, we can de-render it to some set of valid attributes.

We postulated above that there always exists a higher-level grammar with an axiom that includes all the symbols of the observed grammars. We go a step further and view the capsule network as incomplete if there is more than one observed axiom (\cf Figure \ref{fig:DanglingCaps}). There are four possible causes for this:
\begin{enumerate}[leftmargin=1.3cm]
	\item[\textbf{A.1}] A non-activated parent capsule is lacking a route. \\
	\textit{"What existing symbol best describes these parts?"}
	\item[\textbf{A.2}] A parent capsule is missing. \\
	\textit{"What new symbol best describes these parts?"}
	\item[\textbf{B.1}] An attribute is lacking training data. \\
	\textit{"What existing attribute best describes this style or pose?"}
	\item[\textbf{B.2}] An attribute is missing. \\
	\textit{"What new attribute best describes this style or pose?"}
\end{enumerate}

We may remedy these causes using one of two methods, either triggering the creation of a new route in an existing or new capsule (\textbf{(A.1)}, \textbf{(A.2)}) or triggering the training of an existing or new attribute (\textbf{(B.1)}, \textbf{(B.2)}).

However, deciding which of the four causes is responsible for the multiple observed axioms in the current forward pass is subjective even for humans. For example, consider a capsule network that has  $\left[\textit{leg\,}\right]$, $\left[\textit{panel\,}\right]$ and $\left[\textit{chair\,}\right]$ capsules. It encounters a new scene and the observed parse-tree contains four $\left[\textit{leg\,}\right]$ activations and one $\left[\textit{panel\,}\right]$ activation. $\left[\textit{chair\,}\right]$, however, did not activate, triggering the meta-learning pipeline, due to multiple observed axioms. Is this just a $\left[\textit{chair\,}\right]$ with a previously unknown style \textbf{(B.2)}? Or is this a completely new capsule such as $\left[\textit{stool\,}\right]$ \textbf{(A.2)}? 

\begin{table}[ht]
	\begin{center}
		\begin{tabular}{|| c | c | c | c | c ||} 
			\hline
			Feature & A.1 & A.2 & B.1 & B.2 \\
			\hline\hline

			\makecell{Observed Axioms have same $\Omega$ as parent} & \cellcolor{yellow!10!white} 4 & \cellcolor{yellow!10!white} 3 & \cellcolor{green!10!white} 14 & \cellcolor{green!10!white} 12  \\ 
			\hline
			
			\makecell{Observed Axioms don't have same $\Omega$ as parent} & \cellcolor{yellow!10!white} 5 & \cellcolor{green!10!white} 19 & \cellcolor{red!10!white} 1 & \cellcolor{red!10!white} 0  \\ 
			\hline
			
			\makecell{Parts tracked from previous scenes} & \cellcolor{green!10!white} 14 & \cellcolor{red!10!white} 1 & \cellcolor{green!10!white} 17 & \cellcolor{green!10!white} 12 \\ 
			\hline
			
			\makecell{$\Omega: Z(\vec{\alpha}, \vec{\tilde{\alpha}})$ indicates one attribute mismatch \\ with no entry in memory $\alpha^i >\epsilon$ } & \cellcolor{red!10!white} 1 & \cellcolor{red!10!white} 0 & \cellcolor{green!10!white} 12 & \cellcolor{red!10!white} 2  \\ 
			\hline
			
			\makecell{$\Omega: Z(\vec{\alpha}, \vec{\tilde{\alpha}})$ indicates attribute mismatch \\ for (position, rotation, size) only} & \cellcolor{yellow!10!white} 4 & \cellcolor{yellow!10!white} 3 & \cellcolor{green!10!white} 13 & \cellcolor{green!10!white} 10  \\ 
			\hline			
			
			\makecell{$\Omega: Z(\vec{\alpha}, \vec{\tilde{\alpha}})$ indicates attribute mismatch \\ for more than half of all attributes} & \cellcolor{green!10!white} 12 & \cellcolor{green!10!white} 14 & \cellcolor{yellow!10!white} 4 & \cellcolor{yellow!10!white} 4  \\ 
			\hline	
			
			\makecell{$\cdots$} &  &  &  &   \\ 
			\hline

			\hline
			
		\end{tabular}
	\end{center}
	\caption{\label{tab:decisionMatrix} Example of a trained decision matrix with an excerpt of features derived from the observed parse-trees and what cause they indicate (number of past oracle decisions). Here $\Omega$ is the capsule with the highest $p_\Omega$ that did not activate.}
\end{table}

We, thus, introduce a decision matrix (\cf Table \ref{tab:decisionMatrix}). Akin to child development, we train this matrix by querying an oracle in the early stages of the capsule network's training process and update the entries. Decisions are made by summing up all rows of features that evaluate to true and finding the column with maximum value. We may remove the oracle at any point in time, as it does not impair the learning capability of the network itself, only the ability of the agent to make human-like decisions and learn the correct names.

\medskip

\noindent\textbf{Lexical Interpretation.} We interpret our grammar lexically. It is easy to see that each symbol represents a compound noun ($\left[\textit{chair\,}\right]$ or $\left[\textit{dining-room\,}\right]$). For attributes, this analysis is more involved. Note that we have three attributes which we treated differently in Equations \ref{eq:Gamma1}-\ref{eq:Gamma3}: $\vec{\alpha}^{\; rot}, \vec{\alpha}^{\; size}$ and $\vec{\alpha}^{\; pos}$. We interpret these as prepositions. This becomes obvious, once we have multiple objects in a scene and are able to refer to their spatial relationship using words such as "on" or "near", based purely on these attributes.

For static scenes, we interpret all remaining attributes as adjectives, such as "wooden" or "red". Their magnitude is then related to adverbs, such as "very". However, in dynamic scenes, some attributes of an object change over time and describe new poses for the parts. Thus, we interpret these time-dependent attributes as verbs, such as "walk". Their value is equivalent to the normalized time evolution of an animation.

These interpretations are both interesting semantically, as well as for querying the oracle. Instead of presenting a choice between \textbf{(A.1)} - \textbf{(B.2)} and some values, an actual question can be formed from the activated features (\cf Table \ref{tab:decisionMatrix}). Consider a capsule network that has thus far only seen a modern $\left[\textit{chair\,}\right]$, made out of a blend of metal and wood. We now show it a chair made of the same parts, but with less metal and in a classical design. Even though the capsule was trained with basic style transformations $U$, the design is still too complex to grasp. Instead, meta-learning is triggered by cause \textbf{(B.2)}, because of a mismatch of attributes "metallic", "wooden" and "modern" in $Z(\vec{\alpha}, \tilde{\vec{\alpha}})$. As we have access to a lexical interpretation, we can make these abstract pieces of information easier to understand for a human oracle, by letting the meta-learning agent pose an actual question: "This object looks similar to a modern chair, but is very wooden instead. What adjective best describes this style?". The answer "classical" then adds a new attribute to the capsule.

\section{Implementation, Results and Comparison}

\noindent\textbf{Implementation.} We implement the renderer $g$ of primitive capsules using signed distance fields \cite{Quílez:2017} and their encoder $\gamma$ using an AlexNet-like convolutional neural network \cite{Krizhevsky:2012} for regression. For semantic capsules we use Equations \ref{eq:Gamma1}-\ref{eq:Gamma3} for $\gamma$ and a 4-layer deep dense regression neural network with $\tanh$ activation functions for $g$. The training data is generated synthetically using the process described in this paper and all hyperparameters, such as learning rate, are fine-tuned by hand. Our implementation called VividNet is found on Github at \url{https://github.com/Kayzaks/VividNet}. 

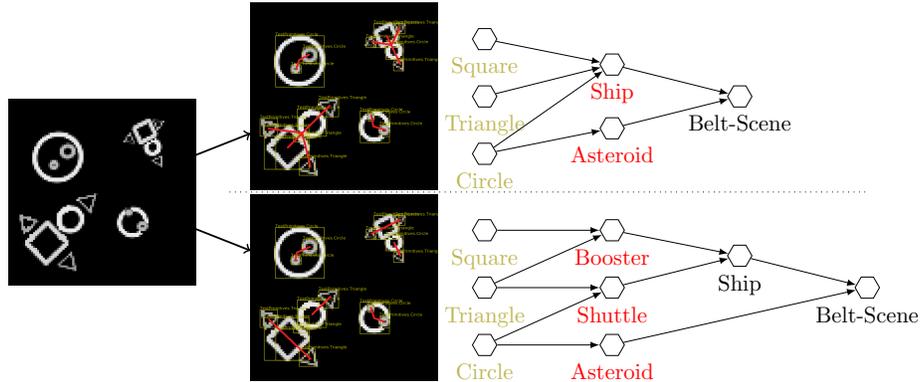
\begin{figure}[ht]
	\centering
	\begin{adjustbox}{max width=1.0\textwidth}
		\begin{tikzpicture}
		\node[inner sep=0pt] (A0) at (0,0) {\includegraphics[width=.24\textwidth]{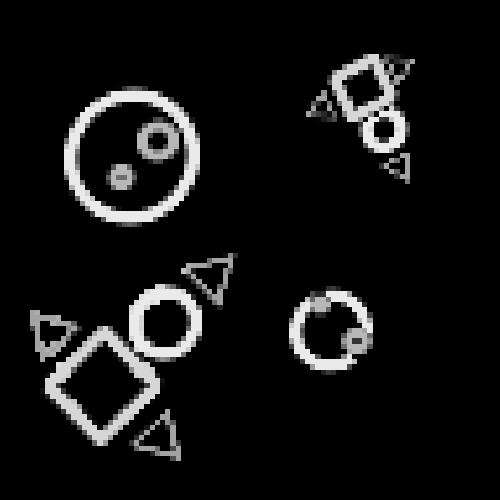}};
		\node[inner sep=0pt] (A2) at (3.8,1.5) {\includegraphics[width=.24\textwidth]{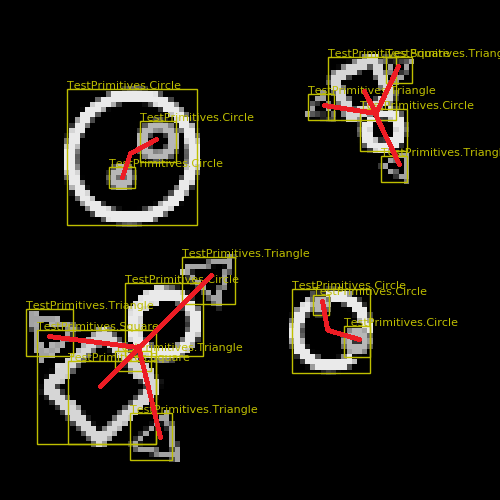}};
		\node[inner sep=0pt] (A3) at (3.8,-1.5) {\includegraphics[width=.24\textwidth]{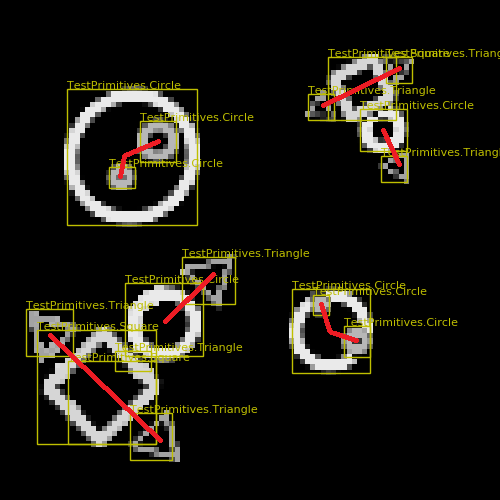}};
		
		\draw[->,thick] (A0) -- (A2);
		\draw[->,thick] (A0) -- (A3);
		
		\node (C01) [label=below:\textcolor{black!30!yellow}{Square}, regular polygon,regular polygon sides=6, draw=black, fill=white] at (6,2.4) {};
		\node (C02) [label=below:\textcolor{black!30!yellow}{Triangle}, regular polygon,regular polygon sides=6, draw=black, fill=white] at (6,1.5) {};
		\node (C03) [label=below:\textcolor{black!30!yellow}{Circle}, regular polygon,regular polygon sides=6, draw=black, fill=white] at (6,0.6) {};
		
		\node (C11) [label=below:\textcolor{red}{Ship}, regular polygon,regular polygon sides=6, draw=black, fill=white] at (8,2.0) {};
		\node (C12) [label=below:\textcolor{red}{Asteroid}, regular polygon,regular polygon sides=6, draw=black, fill=white] at (8,1.0) {};
		
		\node (C21) [label=below:\textcolor{black}{Belt-Scene}, regular polygon,regular polygon sides=6, draw=black, fill=white] at (10,1.5) {};
		
		\draw[tochild, black] (C01) -- (C11);
		\draw[tochild, black] (C02) -- (C11);
		\draw[tochild, black] (C03) -- (C11);
		\draw[tochild, black] (C03) -- (C12);
		
		\draw[tochild, black] (C11) -- (C21);
		\draw[tochild, black] (C12) -- (C21);
		
		\draw [dotted] (2,0) -- (12,0);
		
		\node (D01) [label=below:\textcolor{black!30!yellow}{Square}, regular polygon,regular polygon sides=6, draw=black, fill=white] at (6,-0.6) {};
		\node (D02) [label=below:\textcolor{black!30!yellow}{Triangle}, regular polygon,regular polygon sides=6, draw=black, fill=white] at (6,-1.5) {};
		\node (D03) [label=below:\textcolor{black!30!yellow}{Circle}, regular polygon,regular polygon sides=6, draw=black, fill=white] at (6,-2.4) {};
		
		\node (D11) [label=below:\textcolor{red}{Booster}, regular polygon,regular polygon sides=6, draw=black, fill=white] at (8,-0.6) {};
		\node (D12) [label=below:\textcolor{red}{Shuttle}, regular polygon,regular polygon sides=6, draw=black, fill=white] at (8,-1.5) {};
		\node (D13) [label=below:\textcolor{red}{Asteroid}, regular polygon,regular polygon sides=6, draw=black, fill=white] at (8,-2.4) {};
		
		\node (D21) [label=below:\textcolor{black}{Ship}, regular polygon,regular polygon sides=6, draw=black, fill=white] at (10,-1.0) {};
		
		\node (D31) [label=below:\textcolor{black}{Belt-Scene}, regular polygon,regular polygon sides=6, draw=black, fill=white] at (12,-1.5) {};
		
		\draw[tochild, black] (D01) -- (D11);
		\draw[tochild, black] (D02) -- (D11);
		\draw[tochild, black] (D02) -- (D12);
		\draw[tochild, black] (D03) -- (D12);
		\draw[tochild, black] (D03) -- (D13);
		
		\draw[tochild, black] (D11) -- (D21);
		\draw[tochild, black] (D12) -- (D21);
		
		\draw[tochild, black] (D21) -- (D31);
		\draw[tochild, black] (D13) -- (D31);
		
		\end{tikzpicture}
	\end{adjustbox}
	\caption{Two of many possible capsule network configurations the meta-learning agent might end up with, depending on the oracle and decision matrix.} \label{fig:Results}
\end{figure}

\noindent\textbf{Results.} In the initial phase, our capsule network has three primitive capsules: $\left[\textit{square\,}\right]$, $\left[\textit{triangle\,}\right]$ and $\left[\textit{circle\,}\right]$. We begin by showing it an image of a spaceship (LHS of Figure \ref{fig:Results}), upon which it detects all the relevant graphical primitives, such as three triangles, one circle and one square, but has no semantic understanding of their relation. As this constellation leads to five activated capsules with no common parent, the meta-learning agent is called into action. In this case, it is obvious that \textbf{(A.2)} is triggered, as there are no semantic capsules yet. The exact split, however, is subjective and up to the oracle. We may treat these primitives as one space-ship (top row of Figure \ref{fig:Results}) or group them into two independent parts, booster and shuttle, which make up the space-ship (bottom row of Figure \ref{fig:Results}). In either case, the capsule network is extended by new capsules and trained using only this one example.

Now, the capsule network is shown a new scene, which includes an asteroid made up of three circles. The routes of the $\left[\textit{ship\,}\right]$, $\left[\textit{booster\,}\right]$ and $\left[\textit{shuttle\,}\right]$ capsules find no agreement, as none of them have three circles as their parts. Again, the meta-learning agent queries the oracle, which concludes that a new $\left[\textit{asteroid\,}\right]$ capsule is required \textbf{(A.2)}. The asteroids, however, vary quite a bit. In a new scene with a different asteroid, three circles are detected. This time however, a parent capsule ($\left[\textit{asteroid\,}\right]$) does exist, that admits all three circles as its parts, but due to the different configuration did not activate. This leads to a different set of activated features in our decision matrix and we find, after querying the oracle, that the $\left[\textit{asteroid\,}\right]$ is merely missing a route \textbf{(A.1)}. Alternatively, the agent could have concluded that the capsule is missing a style attribute \textbf{(B.2)}.

In Figure \ref{fig:Results} we show two of the many possible timelines the training process could have taken, depending on the choice of features in the decision matrix, as well as the response by the oracle during the meta-learning process. It was sufficient to show the capsule network one spaceship (or its parts) and a few asteroids to construct the entire network and correctly identify these objects and all their attributes in a new scene. 

\noindent\textbf{Comparison.} Our approach differs too much from current classification methods in order to make a direct numeric comparison. The neural symbolic capsule network can only express confidence, but has no notion of accuracy, as any inaccuracy is remedied by the meta-learning pipeline and its oracle. This does not mean it has perfect accuracy, but rather that it continues to learn forever. Further, the initial choice of primitive capsules is very important in the overall performance of the network. Any comparison would, thus, need to fixate the capsule network in a subjective configuration, eliminating the benefit of lifelong meta-learning.

\section{Conclusion and Outlook}

In this work we showed the internal workings of our neural-symbolic capsule network and how it extends itself through lifelong meta-learning. The proposed network is bi-directional: Feed-forward (\ie, the capsule network) it is a pattern recognition algorithm and feed-backward (\ie, the generative grammar) it is a procedural graphics engine. The ability to render allows us to generate a segmentation mask for all detected objects and their components. However, our reliance on rendering for primitive capsules as the underlying mechanism for inverse graphics also comes with the downside that we are limited by the current state of computer graphics for detection.

We also showed how the network is capable of learning to detect new objects using a few-shot approach and that the training process is very human-like. This allows it to grow indefinitely with less training data, but requires the presence of an oracle to provide nouns, adjectives or verbs for new discoveries, replacing the large amounts of hand labeled data found in the classical approach.

We believe that by next focusing on video data as input and coupling the system with intuitive physics \cite{Battaglia:2016,Hamrick:2017}, we may extend the inverse-graphics capabilities to inverse-simulation.

\pagebreak

\bibliographystyle{splncs04}
\bibliography{egbib}

\end{document}